\documentclass[conference]{IEEEtran}
\newcommand{\eg}{e.g.\ }

\ifCLASSINFOpdf
\else
\fi
\usepackage{xcolor}
\usepackage{url}
\usepackage{hyperref}

\usepackage{latexsym}
\usepackage{amsmath}
\usepackage{amssymb}
\usepackage{amsfonts}
\usepackage{amstext}
\usepackage{cite}
\usepackage[normalem]{ulem}
\usepackage{subcaption}
\usepackage{array}
\usepackage{wrapfig}
\usepackage{multirow}
\usepackage{tabularx}
\usepackage{booktabs}
\usepackage{amsthm}
\usepackage{mdframed}

\theoremstyle{definition}
\newtheorem{definition}{Definition}

\usepackage[export]{adjustbox}

\usepackage{listings}
\usepackage{xcolor}

\definecolor{codegreen}{rgb}{0,0.5,0}
\definecolor{codegray}{rgb}{0.5,0.5,0.5}
\definecolor{codepurple}{rgb}{0.5,0,0.8}
\definecolor{backcolour}{gray}{0.97}

\lstdefinestyle{mystyle}{
    backgroundcolor=\color{backcolour},   
    commentstyle=\color{codegreen},
    keywordstyle=\color{magenta},
    stringstyle=\color{codepurple},
    basicstyle=\ttfamily\footnotesize,
    breakatwhitespace=false,         
    breaklines=true,                 
    captionpos=b,                    
    keepspaces=true,                 
    showspaces=false,                
    showstringspaces=false,
    showtabs=false,                  
    tabsize=2
}

\begin{document}
%
\title{A New Approach to Complex Dynamic Geofencing for Unmanned Aerial Vehicles}


\author{\IEEEauthorblockN{Vihangi Vagal}
\IEEEauthorblockA{\textit{Risk Advisory} \\
\textit{Deloitte LLP} \\
London, United Kingdom \\
vihangiv@gmail.com}
\and
\IEEEauthorblockN{Konstantinos Markantonakis}
\IEEEauthorblockA{\textit{Information Security Group} \\
\textit{Royal Holloway University of London}\\
Egham, United Kingdom \\
k.markantonakis@rhul.ac.uk}
\and
\IEEEauthorblockN{Carlton Shepherd}
\IEEEauthorblockA{\textit{Information Security Group} \\
\textit{Royal Holloway University of London}\\
Egham, United Kingdom \\
carlton.shepherd@rhul.ac.uk}}


%


\maketitle

\begin{abstract}
The anticipated widespread use of unmanned aerial vehicles (UAVs) raises significant safety and security concerns, including trespassing in restricted areas, colliding with other UAVs, and disrupting high-traffic airspaces. To mitigate these risks, geofences have been proposed as one line of defence, which limit UAVs from flying into the perimeters of other UAVs and restricted locations. In this paper, we address the concern that existing geometric geofencing algorithms lack accuracy during the calculation of complex geofences, particularly in dynamic urban environments. We propose a new algorithm based on alpha shapes and Voronoi diagrams, which we integrate into an on-drone framework using an open-source mapping database from OpenStreetMap. To demonstrate its efficacy, we present performance results using Microsoft's AirSim and a low-cost commercial UAV platform in a real-world urban environment.
\end{abstract}


%
\IEEEpeerreviewmaketitle

\section{Introduction}

Geofencing is a widely used security technique for preventing UAVs from flying into controlled airspaces, such as power plants, airports, and military installations. In general, geofences compare the UAV's current geographical location  with predefined or dynamically identified no-fly zones (NFZs) and other restricted areas.  The UAV operator is then notified of any potential or existing contraventions. On some UAV platforms, this may be followed by the termination of the drone's rotors or the automatic triggering of an emergency landing. Geofences are also used to avoid dangerous UAV-to-UAV collisions, particularly in urban airspaces~\cite{Cho,stevens2016multi,stevens2020geofence}.  Numerous commercial and non-commercial geofencing systems have already been developed, such as NASA's Safeguard project~\cite{dill2018safeguard} and DJI's Geospatial Environment Online (GEO)~\cite{DJI} system.  

Existing geofencing algorithms, like polygonal- and circular-based geofencing, enclose restricted regions using geometric shapes/boundaries to which the UAV's current location is compared. This paradigm is inherently reliant on the accuracy of the geometric boundary with respect to a restricted physical area, \eg airport, prison, or school. If the geofence boundary is larger than the physical object in reality, then the UAV may be prevented from entering legitimate locations. In urban environments, this may arise if restricted areas/NFZs are close to residential areas. Conversely, if the geofence is smaller than the physical object, then operators run the risk of mid-air UAV collisions and inadvertently entering NFZs, which is a criminal offence in many jurisdictions. This challenge is most pronounced in dynamic UAV environments, where precise geofences may not easily be determined in advance and must be continuously monitored and re-evaluated.

In light of these issues, we evaluate a new approach to geometric geofencing based on the application of $\alpha$-shapes, proposed by Edelsbrunner et al.~\cite{edelsbrunner1983shape}, and Voronoi diagrams. We integrate our proposal using OpenStreetMap and provide an empirical analysis of existing geofencing techniques using Microsoft's AirSim and a Navio2 UAV in a real-world urban environment in the United Kingdom. This analysis includes experimental results of the accuracy and performance of our proposal against traditional polygonal geofencing.

We anticipate that our proposal will provide a new method for mitigating mid-air collisions, \eg for UAV swarms, and for preventing incursions into restricted airspaces. While the focus of this paper is UAVs, geofences are also used in fleet and freight management, maritime scenarios, and for user authentication~\cite{wawrzyniak2016application,reclus2009geofencing,haofeng2019wi}. Our proposed methods may also be applicable in these domains. In short, our contributions are as follows:

\begin{itemize}
    \item An analytical and experimental comparison of existing geofencing techniques, their accuracy, performance, and shortcomings.
    \item The evaluation of a novel geofencing method based on $\alpha$-shapes and Voronoi diagrams, which utilises an on-drone geofencing database.
    \item Implementation and experimental results of our proposed technique within a simulated environment using Microsoft's AirSim and an urban deployment using a Navio2-based UAV.
\end{itemize}

\section{Geofencing Methods}
\label{GeofencingMethodsSection}

\subsection{Overview and Setup}

Geofences typically fall into one of two paradigms: \emph{static} geofences remain the same overtime over restricted regions, \eg airports and military bases, which are referenced from a database of predefined locations. \emph{Dynamic} geofences are generated in real-time around unforeseen sensitive areas and obstructions on the UAV's flight path; for example, as spheres around other UAVs. Both approaches necessitate up-to-date geofencing databases and reliable location data feeds.

\begin{figure}  
    \centering
    \includegraphics[height=0.55\linewidth,width=\linewidth]{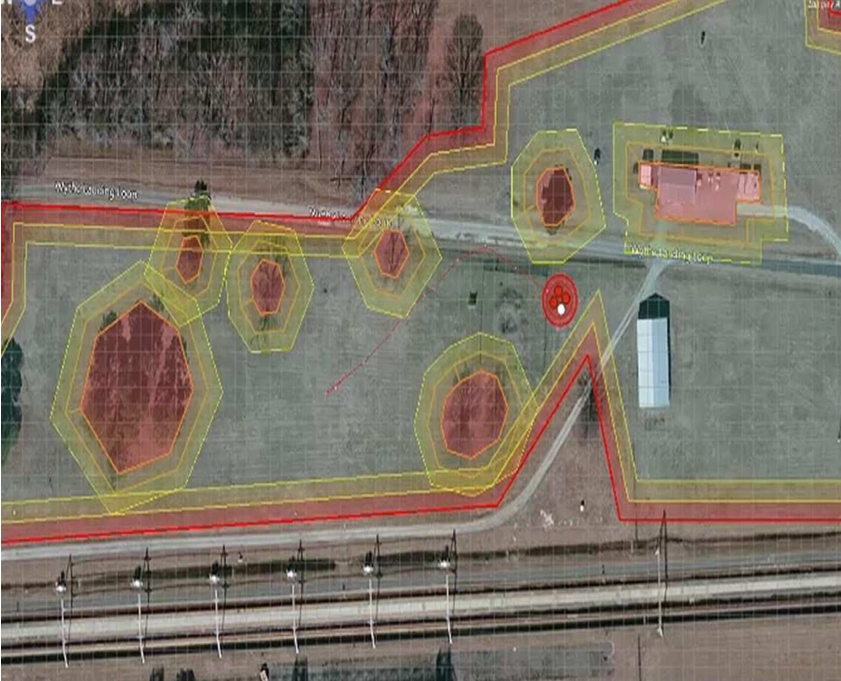}
         \caption{Dual geofencing used by Safeguard~\cite{safeguard:nasa_webpage}, showing NFZ (red), termination (orange) and warning areas (yellow).}
         \label{fig:safeguard}
\end{figure}

Besides methods of generation, geofences can be categorised into three modes of operation. \emph{Keep-in} geofences limit UAVs from flying outside a predefined boundary. An example is NASA's UAS Traffic Management (UTM) system, which uses height, vertical and horizontal buffers for geospatial containment with respect to a cylindrical volume surrounding the UAV~\cite{johnson2017exploration}. \emph{Keep-out} geofences, meanwhile, are formed around restricted areas to prevent UAVs from flying into them. DJI, a leading commercial drone manufacturer, uses keep-out geofences for its UAVs. 
DJI UAVs are restricted to operate in specific geofenced areas, including airports, prisons, and power plants. The system employs an `enhanced warning zone' around NFZs in which drone pilots are notified. If the NFZ is violated, the operator loses control over the UAV before undergoing an immediate landing procedure; the UAV cannot take off or power-up within an NFZ~\cite{DJI}.  Lastly, \emph{dual} geofences use keep-in \emph{and} keep-out methods, which can be beneficial in urban environments~\cite{Cho}. They limit UAVs to a set area from the operator while respecting NFZs within that area. The NASA Safeguard system uses such an approach~\cite{dill2018safeguard,safeguard:nasa_webpage}, which uses multiple warning and UAV termination layers (see Fig.~\ref{fig:safeguard}).

\begin{figure}
    \centering
        \includegraphics[width=0.85\linewidth]{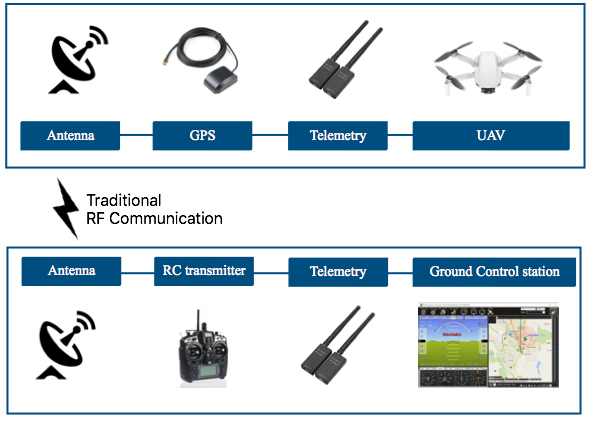}
        \caption{High-level GCS-to-UAV communication architecture.}
          \label{fig:GCS1} 
\end{figure}

In this work, we consider geofencing from the perspective of a UAV that communicates with a ground control station (GCS). The essential components of such a setup are shown in Fig.~\ref{fig:GCS1} and described as follows: 
\begin{itemize}
    \item\textbf{Global navigation satellite system (GNSS)}: Provides the UAV's current location in the form of latitude, longitude and altitude coordinates in real time. GPS, GLONASS, BDS, or Galileo may be used depending on the target deployment region.
    \item\textbf{Ground Control Station (GCS)}: Communicates with the UAV during its operation, which may be used for direct control (manual operation) and for displaying geofencing infringements to the operator.
    \item\textbf{Telemetry module (TM)}: The ground telemetry module is attached to the GCS and the air telemetry module to the drone. The UAV and GCS send and receive essential data using these modules.
    \item\textbf{Radio-frequency transceiver (RF)}: The RF module is part of the GCS; it is coupled with TM and the antenna to communicate with the UAV.
    \item\textbf{Flight controller}: A hardware-firmware setup whereby UAV actuators are synchronously connected and controlled; ArduPilot~\cite{ardupilot}, PX4~\cite{px4}, Cleanflight~\cite{cleanflight}, Navio~\cite{navio2}, and Betaflight~\cite{betaflight} are some widely used flight controllers for commercial UAVs.
    \item\textbf{Geofencing database}:  A database referenced by the UAV that contains the details of restricted locations. The database may include coordinates of those regions and additional restrictions, such as permitted times of operation, noise levels, height, and video restrictions.
\end{itemize}

Today's commercial UAVs are heavily reliant on GNSS measurements for detecting potential and current geofence incursions. This is performed by comparing measurments to those in a geofencing database using a desired geometric geofencing algorithm (see Section \ref{sec:standard_algorithms}), which can be integrated into a GCS or on the UAV itself, including within autopilot modules.  Geofences may be predetermined during the mission planning stage or by dynamically referencing a data source mid-flight. In both cases, geofences may be created with the assistance of open-source or proprietary mapping tools in tandem with human input and/or artificial intelligence. It is important to note that inaccurate and imprecise geofence definitions, GNSS sensor failures, adverse weather conditions, and actuator malfunctions may all lead to incursions into restricted airspaces.

\subsection{Geometric Algorithms}
\label{sec:standard_algorithms}
Geometric algorithms are used to continuously evaluate whether a UAV's GNSS coordinates are in contravention of active geofences in $\mathbb{R}^{2}$ or $\mathbb{R}^{3}$ Euclidean space.\footnote{Two-dimensional geofencing is typically used where the UAV's latitude and longitude are restricted, but not its altitude.}  Common geometric methods include polygonal, spherical, cylandrical, and elliptical geofencing~\cite{Cho,Gurriet2016Jun,stevens2017specification,stevens2020geofence}. 

For a point of interest, $p$, e.g.\ the UAV's current location, polygonal geofencing operates by projecting an infinite ray, $\gamma$, through $p$. $p$ is considered to be within the geofence's perimeter if the number of edges that $\gamma$ intersects is odd on either side of $p$.  In comparison, spherical geofencing computes the absolute distance, $d$, between the centre of the sphere of radius $r$ and a point of interest. A point is considered inside the boundary if $d < r$.  This method has lower computational complexity than polygonal geofencing, but lacks the ability to tightly enclose complex physical locations or objects. Circular geofencing is the two-dimensional analogue of this method. Cylandrical and elliptical geofences function similarly by forming virtual cylinder- and ellipse-based perimeters and testing whether the UAV's coordinates lie within those perimeters. 

The main drawback of non-polygon methods is that the distance between object edges and the drawn boundary may significantly differ for complex, polygon-shaped objects.  These methods make it difficult to precisely enclose restricted areas, particularly when multiple restricted areas are nearby. This is further compounded by the use of buffer spaces for mitigating operational uncertainties, such as adverse weather conditions. In urban environments, it is conceivable that the interference between multiple buffered geofences may unnecessarily impede UAV flight plans and the reachability of destinations. 

\subsection{Evaluated Method}

To address the shortfalls of existing geometric algorithms for complex dynamic environments, we evaluate a novel framework based on the application of alpha shapes~\cite{edelsbrunner1983shape}.

\begin{figure}
        \includegraphics[width=\linewidth]{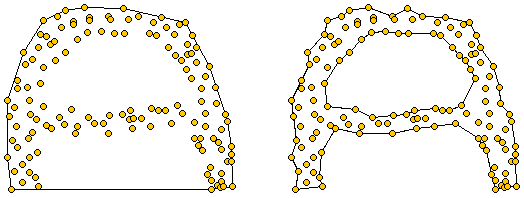}
        \caption{Two $\alpha$-shapes of a discrete set of points in $\mathbb{R}^{2}$ using high (left) and lower (right) values of $\alpha$~\cite{everythingalpha}.}
        \label{fig:alpha_differences} 
\end{figure}
\begin{figure}
\centering
        \includegraphics[width=0.7\linewidth]{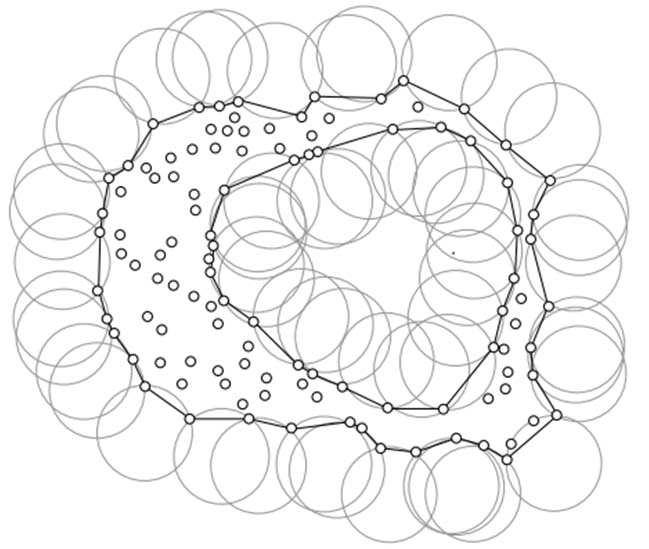}
        \caption{Generating an $\alpha$-shape from disks with radius $\alpha$~\cite{Articque}.}
        \label{fig:generating_alpha} 
\end{figure}

An \emph{alpha} shape, or $\alpha$-shape, of a discrete set of points, $S \in \mathbb{R}^{n}$, is a polytope determined from $S$ and a real value, $\alpha$. It is considered a subgraph of the Delaunay triangulation and a generalisation of the convex hull, which is the intersection of all convex sets containing $S$. When $\alpha \to \infty$, the $\alpha$-shape converges to the convex hull of $S$; as $\alpha \to 0$, the $\alpha$-shape degenerates to the point set $S$. Intuitively, one can tailor this value, $0 \leq \alpha \leq \infty$, to produce an alpha shape graph of varying fineness, as shown in Fig.~\ref{fig:alpha_differences}. Alpha shape graphs have already been shown to be an effective method for generating keep-in and keep-out geofences in complex urban environments~\cite{Cho}. More formally:
\begin{definition}[Alpha shapes~\cite{everythingalpha,edelsbrunner1983shape}] Let a generalised disk of radius, $r$, have the following properties: 
\begin{itemize}
    \item If $\alpha > 0$, it is an ordinary closed disk of radius $r=1/\alpha$.
    \item If $\alpha = 0$, it is a half-plane.
    \item If $\alpha < 0$, it is the complement of a closed disk with $r=-1/\alpha$.
\end{itemize}
Given a set of points, $S$ and a value for $\alpha$, an alpha shape graph is constructed as follows:
\begin{enumerate}
    \item For each point, $p_i \in S$, create a vertex $v_i$.
    \item Create an edge between two vertices $v_i$ and $v_j$, $i \neq j$, when there exists a generalised disk with $r=1/\alpha$ containing $S$ and which satisfies the property that $p_i$ and $p_j$ lies on its boundary (Fig.~\ref{fig:generating_alpha}). 
\end{enumerate}
\end{definition}

\emph{Voronoi diagrams}, also known as Dirichlet tessellations, can also be combined with $\alpha$-shapes. A Voronoi diagram is a tessellation method for partitioning a discrete set of points, $S$, as such:

\begin{definition}[Voronoi diagram] 
Let $d$ be a distance function, such as the Euclidean distance, between two points in a finite set, $(p_i, p_j) \in S$. 
The collection of all points closest to $p_i \in S$ is known as the Voronoi region, $V_{S}(p_i)$, for a metric space, $\mathcal{X}$, e.g. $\mathbb{R}^{3}$: $V_{S}(p_i) = \{ x \in \mathcal{X} \; | \; d(x, p_i) \leq d(x, p_j) \; \forall \; i \neq j$\}. The Voronoi diagram, $\textbf{V}(S)$, is then defined as $\textbf{V}(S) = \{V_{S}(p_1), V_{S}(p_2), \dots, V_{S}(p_n)\}$.
\end{definition}

\begin{figure}
\centering
        \includegraphics[width=0.75\linewidth]{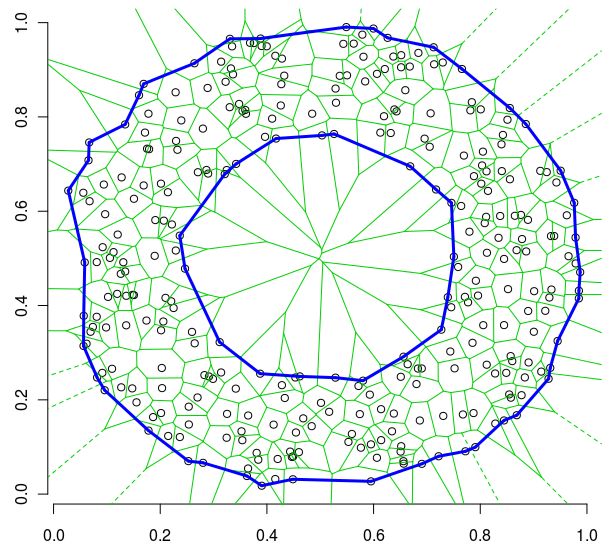}
        \caption{Combining an $\alpha$-shape and a Voronoi diagram to form geofencing boundaries (blue) with internal edges for facilitating UAV path planning~\cite{jordi2012alpha}.}
        \label{fig:combining_alpha}
\end{figure}

Voronoi diagrams have been proposed in the UAV path planning literature as an efficient method for circumnavigating obstructions and restricted areas in dynamic environments~\cite{chen2014uav,pehlivanoglu2012new,scott}. It is possible to use weighted Voronoi graphs in which weights are allocated to graph edges to inform the difficulty of traversing challenging areas~\cite{chen2014uav}. The shortest path between two points can then be deduced using a path finding method, such as A* search or Dijkstra's algorithm. In this work, we propose a combination of Voronoi diagrams and $\alpha$-shapes, illustrated in Fig.~\ref{fig:combining_alpha}, for unifying the benefits of $\alpha$-shape geofencing and the path planning applications of Voronoi diagrams.

\section{System Workflow}
\label{sec:system_design}

\begin{figure}
\centering
\includegraphics[width=\linewidth,frame]{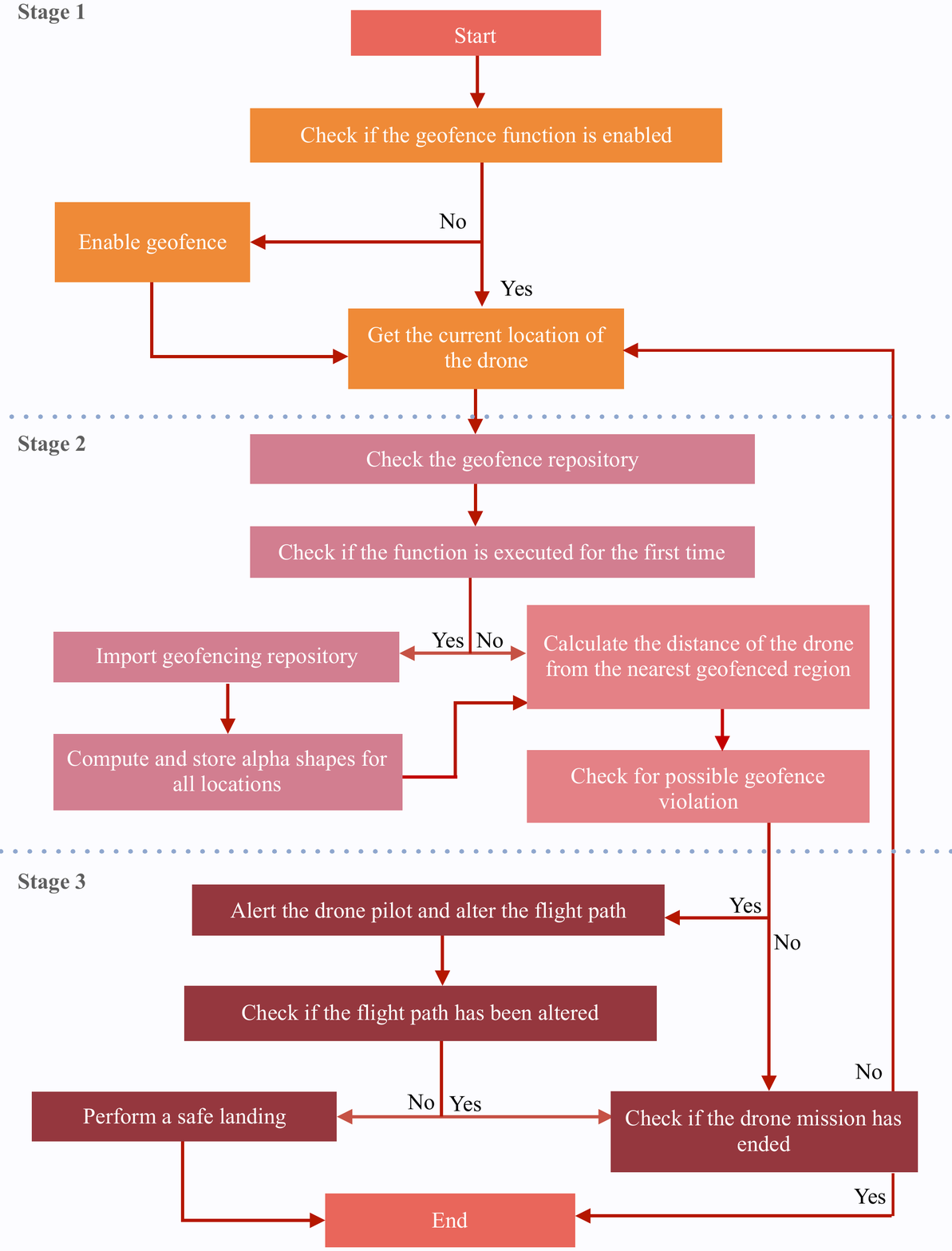}
\caption{System flowchart.}
\label{fig:DG} 
\end{figure}

From a UAV system-level perspective, the flowchart of our geofencing method contains three main stages shown in Fig.~\ref{fig:DG}. The first stage is an initialisation step for enforcing the activation of the UAV's geofencing functionality and ascertaining its current location. Next, in stage two, the platform loads the geofencing repository, i.e.\ the list of NFZs and restricted airspaces, and computes the $\alpha$-shapes corresponding to these locations on the UAV's flight path. The coordinates of the $\alpha$-shapes are then stored for future reference; the pre-computation of $\alpha$-shapes before a take-off allows for faster evaluation and lower response times mid-flight. This repository can be acquired in an online fashion or, in offline environments, the UAV may receive a recent copy of the geographic operating environment pre-flight. The UAV's current location is then compared to these shapes to detect current and potential geofencing violations. Lastly, stage three alerts the UAV operator if a geofencing violation is detected; if needed, the UAV flight controller may terminate the rotors or initiate an emergency landing.

\section{Implementation}

\subsection{AirSim Experiments}

\begin{figure}
    \centering
    \includegraphics[width=\linewidth]{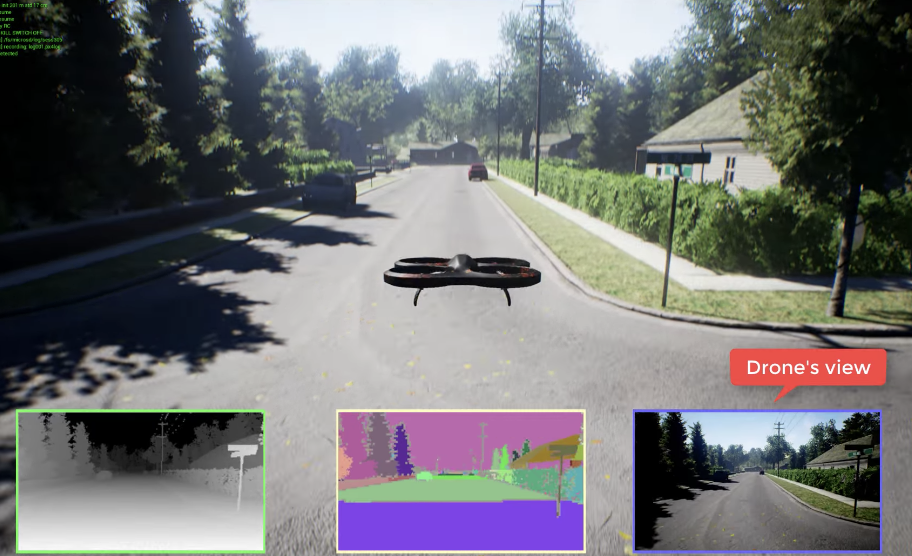}
    \caption{AirSim UAV simulation platform~\cite{shah2018airsim}.}
    \label{fig:airsim}
\end{figure}

AirSim is a cross-platform, open-source simulator developed by Microsoft for autonomous vehicle research based on the Unreal Engine~\cite{shah2018airsim}. It supports software-in-the-loop simulation with widely used, off-the-shelf flight controllers, such as the PX4 and ArduPilot, and hardware-in-loop with PX4 within a virtual environment (Fig.~\ref{fig:airsim}). AirSim provides access to C++ and Python APIs for vehicle control and to retrieve continuous information about the target vehicle. We used these to implement and trial our proposed geofencing method, including its comparison to existing geofencing methods, prior to a real-world deployment. The simulator uses the NED coordinate system, i.e.\ X (horizontal movement), Y (vertical moment), Z (altitude) coordinates, which was used as the basis for determining the UAV's location within the aforementioned workflow in Fig.~\ref{fig:DG}. 
The AirSim built-in GPS module was used to ascertain the UAV's current location.

The algorithm first computes the nearest objects to the UAV using the pre-computed database of objects in the virtual environment.  In our implementation, this was stored as a JSON file, which was parsed before computing and saving the $\alpha$-shapes for the current operating region. For the computation of the $\alpha$-shapes themselves, we leveraged the \texttt{alphashapes} Python package~\cite{alphashape}.  Using this, we instrumented the simulation platform to implement a keep-out geofencing around restricted objects. Moreover, we implemented keep-in geofencing by limiting the UAV to a particular geographic area within the virtual environment. 

The UAV's flight path was determined from the user-generated inputs through the Visual Studio command prompt. The path heading was automatically re-adjusted when there is a possibility of entering a restricted location; if this was overridden by the user, then the UAV undergoes an emergency landing before it enters that area.

\subsection{Navio2 Drone Platform} 
\label{navio2}

\begin{figure}
    \centering
            \includegraphics[width=0.9\linewidth]{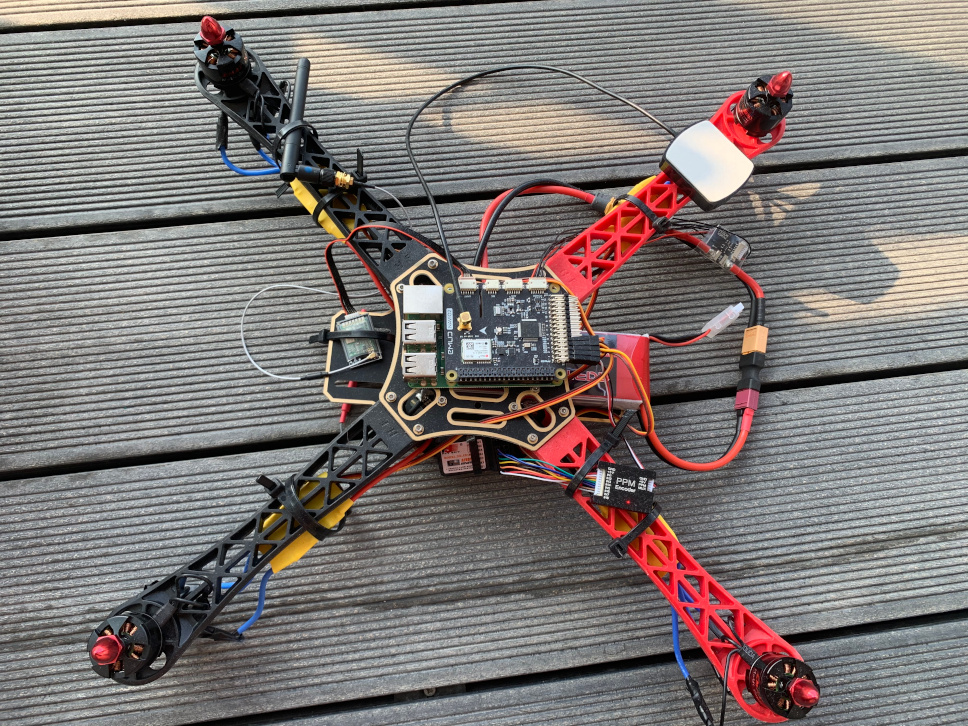}
            \caption{Assembled Navio2-based quadcopter.}
            \label{fig:droneassembly} 
     \end{figure}

After prototyping the proposed method in AirSim, we then evaluated its performance in an urban environment using an off-the-shelf, low-cost ($<$\$500 USD) commercial drone platform. This made use of a Navio2 add-on shield~\cite{navio2}---an autopilot system for the widely used Raspberry Pi board~\cite{RaspberryPi3}---to implement a quadcopter-based UAV (Fig.~\ref{fig:droneassembly}). The components for implementing the hardware platform are as follows:
\begin{itemize}
    \item\textbf{Raspberry Pi 3B+}: A single-board computer with a Broadcom BCM2837B0 system-on-chip (with a quad-core ARM Cortex-A53 at 1.4GHz), 1GB DDR2 SRAM, dual-band wireless LAN, Bluetooth 4.2/BLE, and a 40-pin general-purpose input/output (GPIO) interface~\cite{RaspberryPi3}. 
    \item\textbf{Navio2 Shield}: A Raspberry Pi add-on shield that provides a GNSS receiver---supporting GPS, GLONASS, Galileo and BDS---dual IMUs, a remote-controlled I/O co-processor with 14 PWM output channels for motors and servos, and a high-resolution barometer. The Navio2 supports the Ardupilot open-source autopilot firmware and exposes Python APIs for software-based control; its average current draw is 150mA during operation.
    
    \item\textbf{Radio Controller}: A six-channel transceiver for enabling the UAV to be controlled manually by the user.

    \item\textbf{Telemetry}: A pair of air and ground modules attached to the drone and ground control station. The telemetry modules communicate in-flight data, e.g.\ position and speed, wirelessly between the UAV and GCS. 

    \item\textbf{Battery}: A lithium polymer (LiPo) battery supplies power to the drone with a 3000mAh capacity.
    
    \item\textbf{Actuators}: A combination of motors, propellers and ESCs were selected based on the thrust-to-weight ratio; specifically, 30A brushless motors with 20V ESC and 1048 propellers were used. (ESCs are only required when using brushless motors; brushed motors only need a direct current voltage input). 
    \item\textbf{MavProxy}: The MAVProxy is Ardupilot's GCS implementation a UAVs. It provides a portable GCS for UAVs that support the MAVlink Protocol~\cite{ErleRoboDocs}.
    \item\textbf{Mission Planner}: The Ardupilot mission (APM) planner is an open-source ground station application for MAVlink based autopilots~\cite{ardupilot}. It assists in mission planning using GPS way-points and control events. 
\end{itemize}

\subsection{Geofencing Database}
We used OpenStreetMap as the mapping source for underpinning our geofencing database. OpenStreetMap is an open-source database that supports WGS-84 coordinates practised by many GNSS units \cite{openstreet}. Geographical information about supported countries, e.g.\ USA and UK, can be exported in OSM format from the OpenStreetMap website. This data can be imported into a geographic information system application, like QGIS~\cite{QGIS}, for exporting maps into alternative file formats and coordinate systems, i.e.\ WGS-84 and OSGB-36. 

\begin{figure}
 
        \centering
        \includegraphics[width=\linewidth]{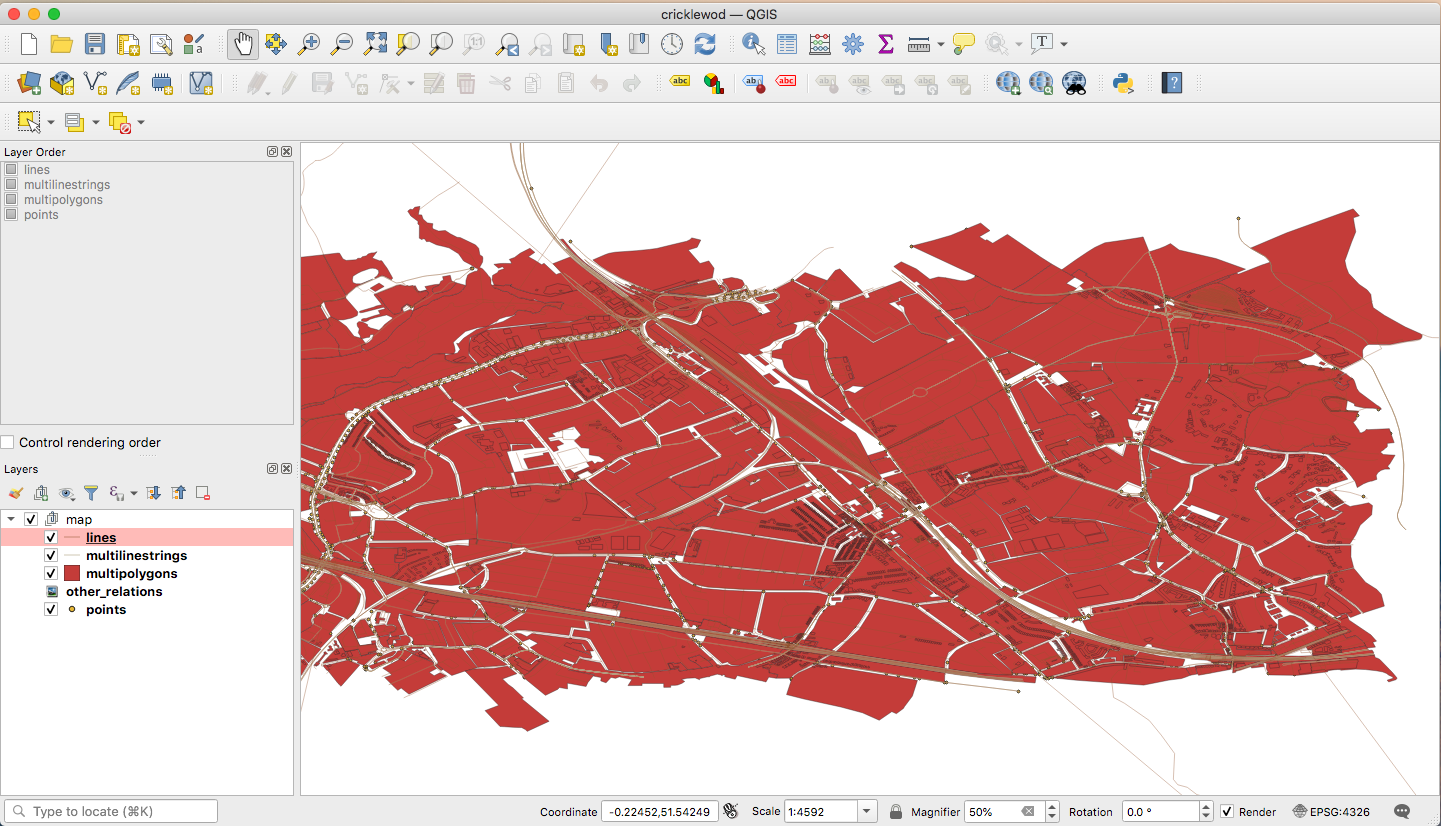}
        \caption{QGIS visualisation.}
         \label{fig: open3} 
          \end{figure}

For our experiments, we imported geographic information about our testing environment, located in the south east region of the United Kingdom, into QGIS (Fig.~\ref{fig: open3}) which was subsequently exported to JSON format. This information  contains features about notable physical locations, e.g.\ location type (military, airport, leisure etc.), its name, short description, and coordinates.  Listing~\ref{fig: open5} shows a sample JSON entry.  The features and coordinates were extracted from the file in Python for constructing the $\alpha$-shapes for each restricted location.

\begin{lstlisting}[language=Python, caption=Sample JSON from OpenStreetMap., label=fig: open5]
{`type': `Feature', `properties': { `osm_id': `533025', `osm_way_id': null, `name': `Canada Copse', `type': `multipolygon', `aeroway': null, `amenity': null, `admin_level': null, `barrier': null, `boundary': null, `building': null, `craft': null, `geological': null, `historic': null, `land_area': null, `landuse': `forest', `leisure': null, `man_made': null, `military': null, `natural': null, `office': null, `place': null, `shop': null, `sport': null, `tourism': null, `other_tags': null }, `geometry': { `type': `MultiPolygon', `coordinates': --truncated--
\end{lstlisting}

We note that, while OpenStreetMap is a widely used geographic database, it has several shortcomings. Its development is community-driven and may lack the maintenance and support of proprietary solutions. It also lacks information about ambient noise, pollution, altitude, and other environmental factors that may practically limit the operation of UAVs.

\section{Evaluation}
\label{evaluationsection}

\begin{figure}
  \centering
  \begin{minipage}[b]{0.48\linewidth}
 
    \includegraphics[width=\linewidth]{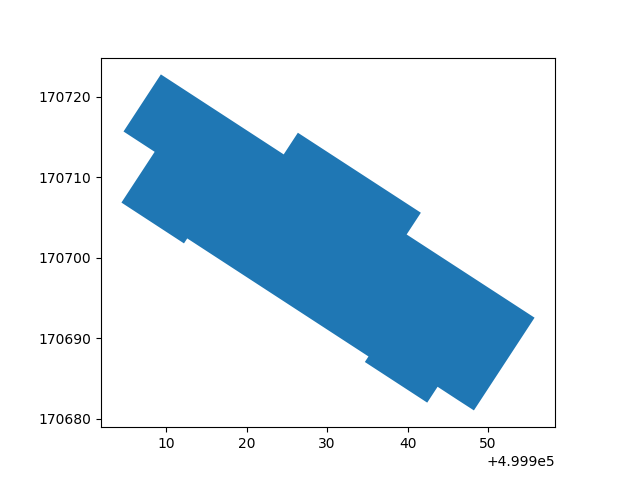}
    \caption{Polygonal geofence.}
    \label{fig:simpoly1} 
  \end{minipage}
  \begin{minipage}[b]{0.48\linewidth}
      \includegraphics[width=\linewidth]{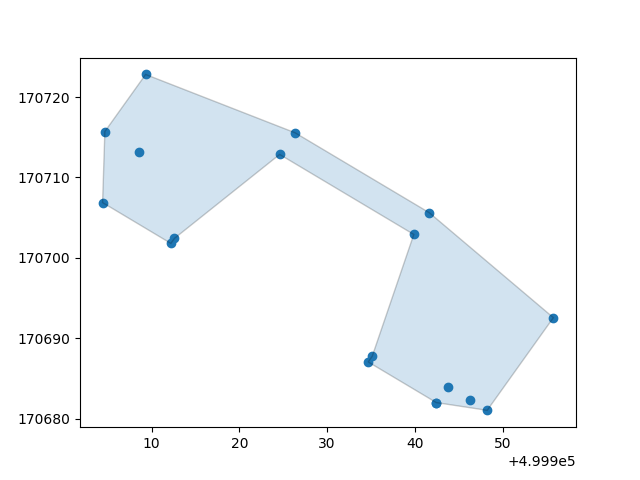}
       \caption{Proposed method.}
        \label{fig:simalpha1} 
  \end{minipage}
\end{figure}

After computing 1,309 geofences in the urban environment, it was revealed that the proposed method consistently and more tightly enclosed physical locations compared to dynamically drawn polygonal geofences (i.e.\ not with human input).
Fig.~\ref{fig:simpoly1} shows an example polygonal geofence, which inaccurately enclosed a far greater physical area than our proposed method shown in Fig.~\ref{fig:simalpha1}.   This process involved labour-intensive human assessment to evaluate the accuracy of both algorithms with respect to ground truth location data.  The initial results are promising, but further experiments across varying geographic regions are warranted to reliably calculate the accuracy of our method.

\begin{figure}
        \includegraphics[width=\linewidth]{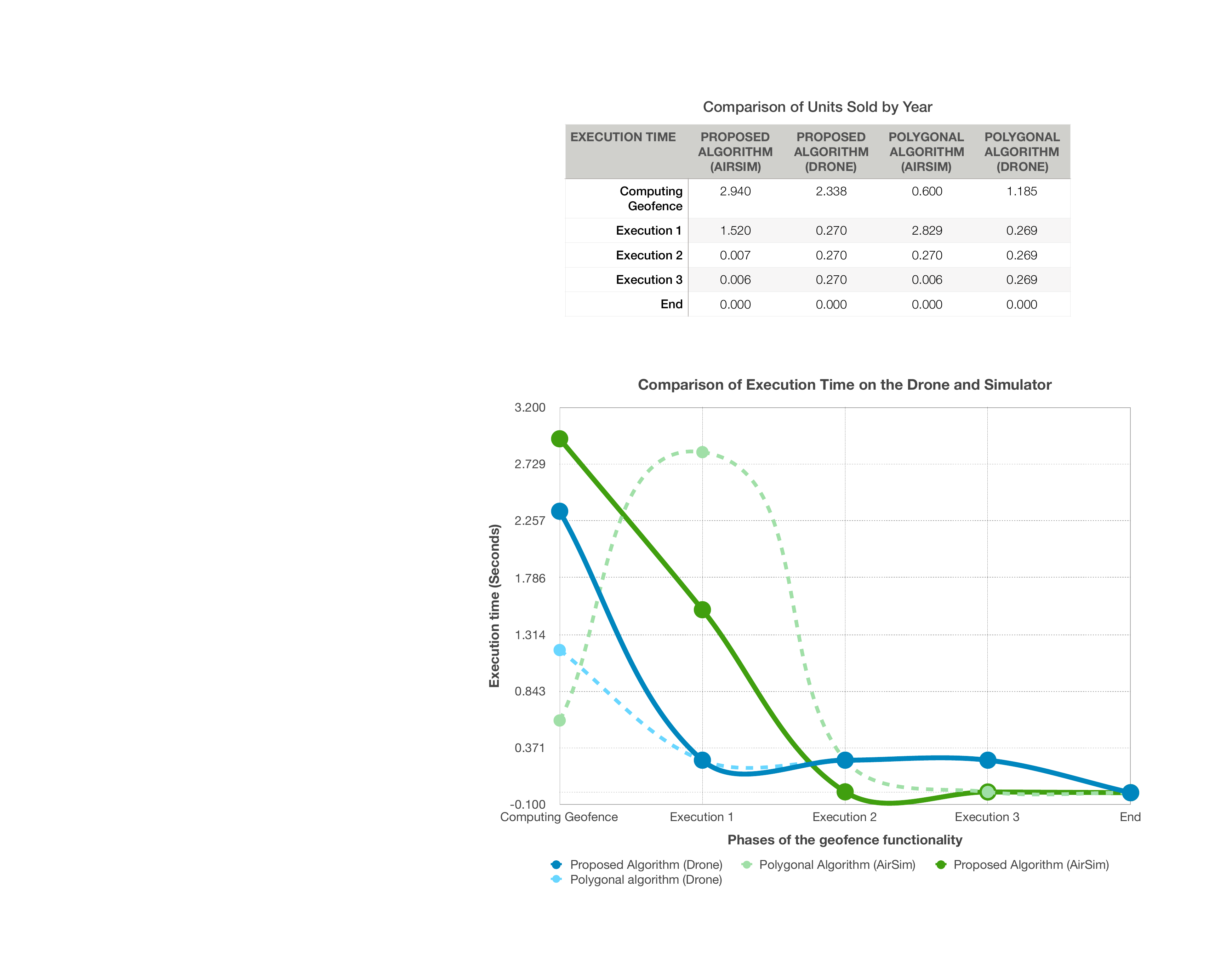}
        \caption{Execution time of polygonal geofencing versus our proposed method on AirSim and the Navio2 UAV platform.}
        \label{fig:simdrone} 
\end{figure}

The computational performance of both approaches was also evaluated at each workflow stage from from Section~\ref{sec:system_design}. Fig. \ref{fig:simdrone} shows the execution time of the proposal versus polygon-based geofencing using AirSim and our physical UAV platform at each stage (on the x-axis).
After the initial computation of the geofences, the execution time for computing the geofences declines to under 500ms on our physical platform. Generally, the polygonal and proposed methods execute within the same order of magnitude---under three seconds for all phases---during operation. However, on average, the proposed method incurs an approximately one second overhead for geofence computation using the Navio2 UAV. Table \ref{tab: algocompare} presents a comparison of the algorithms based on time and the evaluation platform.  Compared to polygonal geofencing, our method requires an additional 1--2 seconds on both AirSim and the physical UAV to generated the geofenced areas. The average run-time execution time for detecting geofence violations is broadly the same: for AirSim, this was 0.006s \emph{vs.} 0.006s (polygonal \emph{vs.} proposed respectively) and, for Navio2, this was 0.269s \emph{vs.} 0.270s.

\begin{table}[ht!]
\centering
\begin{tabular}{|l|l|l|}
\hline
\textbf{\begin{tabular}[c]{@{}l@{}}Comparison\\Stage  Platform \end{tabular}} & \textbf{\begin{tabular}[c]{@{}l@{}}Polygonal \\ Algorithm\end{tabular}} & \textbf{\begin{tabular}[c]{@{}l@{}}Proposed\\  Algorithm\end{tabular}}  \\ \hline
\begin{tabular}[c]{@{}l@{}}AirSim geofence\\computation\end{tabular}              & 0.698 seconds                                                           & 3.44 seconds                                                            \\ \hline
\begin{tabular}[c]{@{}l@{}}AirSim average\\detection time\end{tabular}   & 0.006 seconds                                                           & 0.006 seconds                                                           \\ \hline
\begin{tabular}[c]{@{}l@{}}Navio2 geofence\\computation\end{tabular}        & 1.185 seconds                                                           & 2.338 seconds                                                           \\ \hline
\begin{tabular}[c]{@{}l@{}}Navio2 average\\detection time\end{tabular}    & 0.269 seconds                                                           & 0.270 seconds                                                           \\ \hline

\end{tabular}
\caption{Average performance comparison.}
\label{tab: algocompare}
\end{table}

In summary, the results demonstrate that our proposal has more reliable results than polygon geofencing in relation to geofencing accuracy. This comes with the drawback of a moderate geofence computation overhead, which can be partially mitigated by determining static $\alpha$-shape geofences before the flight commences. 
In general, our investigations suggest that the proposed algorithm provides greater accuracy for complex geofences without a dramatic performance impact in practice. 

\section{Conclusion}
Existing geofencing algorithms can struggle with precisely enclosing restricted airspaces in dynamic environments; for example, when generated in-flight. This can pose safety and accessibility issues in complex urban areas with an abundance of closely located restricted locations, e.g.\ airports, schools and prisons, and permitted airspaces. To address this, we designed and evaluated a new geofencing solution for such scenarios using $\alpha$-shapes and Voronoi diagrams.  We implemented our proposal in a simulated and real-world urban environment using a low-cost, commercially available drone platform. Our results suggest that greater geofencing precision can be achieved while retaining computational performance in the same order of magnitude as polygonal geofencing. We hope that our work provides UAV system designers with a useful option for generating accurate geofences in complex and dynamic operating environments.




\bibliographystyle{IEEEtran}
\bibliography{IEEEabrv,bibliography}
%


\end{document}